\pdfoutput=1

\documentclass[11pt]{article}

\usepackage[final]{acl}

\usepackage{times}
\usepackage{latexsym}

\usepackage{listings}

\usepackage[T1]{fontenc}

\usepackage[utf8]{inputenc}

\usepackage{microtype}

\usepackage{inconsolata}

\usepackage{graphicx}
\usepackage{enumitem}
\usepackage{float}
\usepackage{booktabs} 
\usepackage{multirow}
\usepackage{cellspace}
\usepackage{tabularx}
\usepackage{xspace}
\usepackage{colortbl}
\usepackage{mdframed}
\usepackage{tcolorbox}
\usepackage{longtable}
\usepackage{amssymb}
\usepackage{amsbsy}
\usepackage{bbding}
\usepackage{pifont}
\usepackage{caption}
\usepackage{subcaption}
\usepackage{array}
\usepackage{makecell}
\usepackage{amsmath}

\newcommand{\Ours}{Physics Reasoner}
\newcommand{\revise}[1]{\textcolor{black}{#1}}

\title{Physics Reasoner: Knowledge-Augmented Reasoning \\for Solving Physics Problems with Large Language Models}

\author{
Xinyu Pang\textsuperscript{1}, 
Ruixin Hong\textsuperscript{1}, 
Zhanke Zhou\textsuperscript{2},
Fangrui Lv\textsuperscript{1}, \\
{\bf Xinwei Yang\textsuperscript{1},
Zhilong Liang\textsuperscript{1},
Bo Han\textsuperscript{2}, 
Changshui Zhang\textsuperscript{1}} \\
\textsuperscript{1}Institute for Artificial Intelligence, Tsinghua University (THUAI); \\
\textsuperscript{1}Beijing National Research Center for Information Science and Technology (BNRist); \\
\textsuperscript{1}Department of Automation, Tsinghua University, Beijing, P.R.China \\
\textsuperscript{2}TMLR Group, Hong Kong Baptist University \\
\texttt{\{pangxy22, hrx20, lvfr23, yangxw21, liangzl20\}@mails.tsinghua.edu.cn}, \\
\texttt{\{cszkzhou, bhanml\}@comp.hkbu.edu.hk,} \\
\texttt{zcs@mail.tsinghua.edu.cn} \\
}

\begin{document}
\maketitle

\begin{abstract}
Physics problems constitute a significant aspect of reasoning, necessitating complicated reasoning ability and abundant physics knowledge.
However, existing large language models (LLMs) frequently fail due to a lack of knowledge or incorrect knowledge application.
To mitigate these issues, we propose \Ours{}, a knowledge-augmented framework to solve physics problems with LLMs.
Specifically, the proposed framework constructs a comprehensive formula set to provide explicit physics knowledge and utilizes checklists containing detailed instructions to guide effective knowledge application.
Namely, given a physics problem, \Ours{} solves it through three stages: problem analysis, formula retrieval, and guided reasoning. 
During the process, checklists are employed to enhance LLMs' self-improvement in the analysis and reasoning stages. 
Empirically,
\Ours{} mitigates the issues of insufficient knowledge and incorrect application, achieving state-of-the-art performance on SciBench with an average accuracy improvement of 5.8\%.\footnote{The code is
 publicly available at \url{https://github.com/Xinyu-Pang/Physics_Reasoner}}
\end{abstract}

\section{Introduction}
\begin{figure}[t!]
  \centering
  \includegraphics[width=\columnwidth]{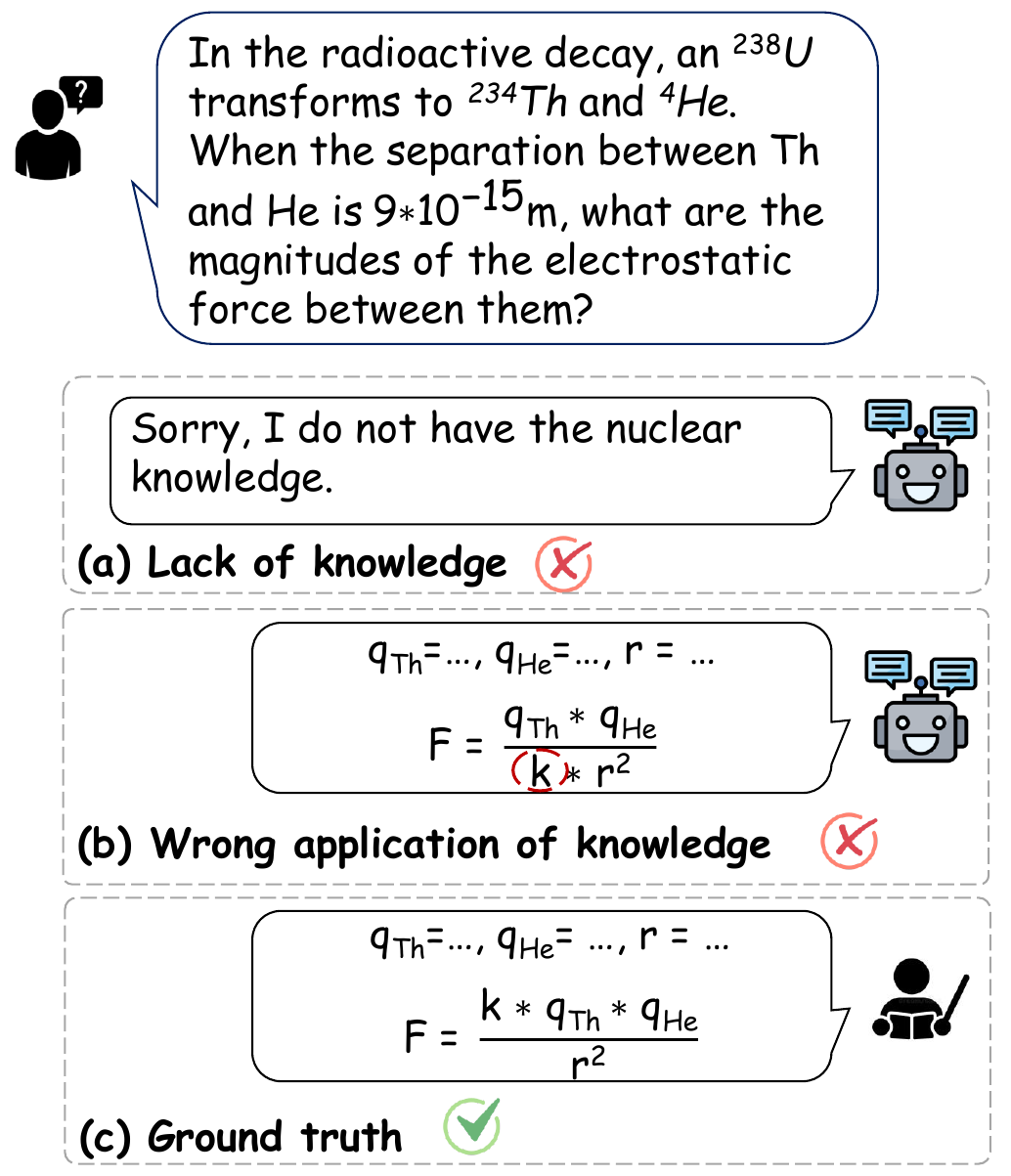}
  \caption{Exemplar error cases. Given a complex physics problem, LLMs may make mistakes due to a lack of physics knowledge, as illustrated in (a), or incorrect application of knowledge, as shown in (b). The ground truth answer to this question is shown in (c).
  }
  \label{fig:intro}
\end{figure}




Physics problems are essential for evaluating the capabilities of large language models (LLMs), assessing the models' comprehension of the natural world, and their ability to navigate complex scenarios. Inadequate performance on physics problems by LLMs could imply a potential deficiency in their understanding of fundamental real-world concepts such as spatial relationships~\cite{DBLP:journals/corr/abs-2311-16714} and molecular structures~\cite{DBLP:journals/corr/abs-2112-03041}, significantly hindering their effectiveness in real-world applications and scientific discoveries.

However, solving physics problems is highly challenging, demanding a high level of abstraction to translate given information into solvable expressions. This process necessitates the acquisition and application of physics knowledge to find an accurate solution \textit{w.r.t.} a given question. 

Several methods are proposed to enhance LLMs' reasoning on physics tasks~\cite{DBLP:journals/corr/abs-2402-12692, DBLP:conf/emnlp/ChenYKLWMXWX23}.
For instance, Chain-of-Thought (CoT)~\cite{DBLP:conf/nips/Wei0SBIXCLZ22} asks models to solve step by step, while Chameleon~\cite{DBLP:conf/nips/LuPCGCWZG23} integrates various external tools.
However, even equipped with advanced methods, 
LLMs still struggle to solve physics problems.
Using the CoT strategy, GPT-3.5-turbo achieves only 6.8\% accuracy on the physics section of SciBench~\cite{DBLP:journals/corr/abs-2307-10635} benchmark.
Considering the knowledge-intensive nature of these physics problems, we raise the research question:
\textit{Can LLMs master physics knowledge and solve physics problems?}

To answer this question, this work starts by analyzing the common errors, which reveal two primary factors behind the frequent failures.
Namely, (1) LLMs lack the necessary physics knowledge, including key concepts and formulae, as illustrated in Fig.~\ref{fig:intro}(a);
and (2) even when provided with relevant knowledge, LLMs still struggle to apply it to solve problems correctly, as shown in Fig.~\ref{fig:intro}(b).

To mitigate these problems, we propose \Ours{}, a novel framework for physics problem reasoning.
Conceptually, this framework aims to make up for the knowledge deficiency of LLMs by incorporating                                                                                                          
(i) a comprehensive formula set for knowledge acquisition, which encompasses $122$ formulae, each accompanied by detailed annotations;
(ii) detailed checklists for evaluating the correctness of \textit{knowledge application}, which contain detailed instructions for identifying and correcting errors, thereby assisting LLMs in effectively applying the acquired knowledge. Furthermore, Python code is integrated into the solving process to guarantee precise and reliable calculations.


Technically, \Ours{} consists of three stages: problem analysis, formula retrieval, and guided reasoning. 
Given a physics problem, it first comprehends the problem and extracts known variables. After the initial extraction, a checklist is used to review the correctness and completeness of the extraction. Next, it retrieves relevant formulae from a pre-constructed formula set and performs reasoning. Finally, it reviews and refines the reasoning process with another checklist to aid LLMs in applying the retrieved knowledge effectively.

To verify the effectiveness of the proposed framework, we conduct extensive experiments on the SciBench benchmark.
Notably, \Ours{} outperforms existing methods by an average of 5.8\% in accuracy, reduces reasoning errors, and further boosts LLMs' physical reasoning abilities.

\vspace{6mm}
Our contributions are summarized as follows:
\vspace{-2mm}
\begin{itemize}[leftmargin=*]
\item We identify the two major limitations of solving physics problems, \textit{i.e.}, (i) lack of knowledge and (ii) incorrect application of knowledge (Sec.~\ref{sec: limitation}).
\vspace{-1mm}
\item We propose a novel reasoning framework that integrates knowledge acquisition with the formula set and application with checklists (Sec.~\ref{sec: approach}).
\vspace{-1mm}
\item We conduct extensive experiments on four datasets. The \Ours{} achieves an average improvement of 5.8\% in accuracy (Sec.~\ref{sec: experiments}).
\end{itemize}

\section{Related work}
\subsection{LLMs for Scientific Reasoning}
The rapid progress of LLMs has significantly advanced the field of scientific reasoning. Numerous representative benchmarks have emerged to evaluate scientific reasoning abilities across a wide array of subjects, including math~\cite{zhou2024can, DBLP:journals/corr/abs-2110-14168, DBLP:journals/corr/abs-2310-02255}, physics~\cite{DBLP:conf/nips/BakhtinM0GG19, DBLP:conf/emnlp/WangDFS23}, and chemistry~\cite{DBLP:conf/nips/GuoGNLGCW023, DBLP:conf/nips/LuMX0CZTCK22}.

LLMs have demonstrated strong abilities in solving scientific problems. 
Recent approaches that gather annotation data and fine-tune LLMs have yielded remarkable results~\cite{DBLP:conf/iclr/Lu0CWZRCK23, DBLP:conf/nips/LewkowyczADDMRS22, DBLP:journals/corr/abs-2402-06852}. 
Also, strategies utilizing prompts with frozen LLMs have proven effective, significantly reducing training overhead while maintaining or even enhancing performance. 
For instance,~\citet{DBLP:conf/iclr/0002WSLCNCZ23} proposes self-consistency, which combines different CoT reasoning paths for more probable answers.

However, despite these advancements,
few methods are specifically designed to address LLMs' knowledge deficiency and their inability to effectively apply the knowledge. 
This gap underscores the need for dedicated approaches to enhance the effectiveness of solving physics-related challenges.

\subsection{Physics Problem Reasoning}
To evaluate LLMs' physical reasoning abilities, several benchmarks have been proposed~\cite{DBLP:conf/nips/BakhtinM0GG19, DBLP:conf/aaai/BiskZLGC20}.
For instance, NEWTON~\cite{DBLP:conf/emnlp/WangDFS23} consists of labeled object-centric data encompassing 8 physical attributes.

However, these works primarily focus on simple physical attributes of everyday objects, neglecting complex physics problem reasoning. Solving complicated physics problems requires sophisticated reasoning skills with the acquisition and application of extensive physics knowledge.

\subsection{Self-improvement}
A widely accepted way to enhance LLMs' abilities is self-improvement through verification feedback, which enhances their capabilities in multiple aspects, including security~\cite{cao2024envisioning, li2023deepinception}, reasoning~\cite{DBLP:journals/corr/abs-2311-07954}, and other critical areas. For example, high-quality feedback can be used for model fine-tuning~\cite{DBLP:conf/nips/Ouyang0JAWMZASR22, DBLP:journals/corr/abs-2303-16749, DBLP:conf/nips/ZelikmanWMG22}. Besides, Feedback can be used to re-rank model outputs~\cite{DBLP:journals/corr/abs-2305-20050, DBLP:journals/corr/abs-2301-00303, DBLP:conf/icml/Ni0RSYWL23}, to modify current reasoning process~\cite{DBLP:conf/nips/ShinnCGNY23, DBLP:journals/corr/abs-2306-03856}, and to guide generation~\cite{DBLP:conf/nips/YaoYZS00N23, DBLP:journals/corr/abs-2305-00633}.
Furthermore,~\citet{DBLP:conf/ijcai/RibeiroWG021} proposes to examine the capacity of LLMs using a checklist, including general capabilities and test types.

However, for complex problems like physics problems, general self-improvement methods often perform poorly and may even have adverse effects, leading LLMs to incorrect reasoning processes.

\section{Limitations of Physics Reasoning}
\label{sec: limitation}
\begin{figure}[t!]
  \centering
  \includegraphics[width=\columnwidth]{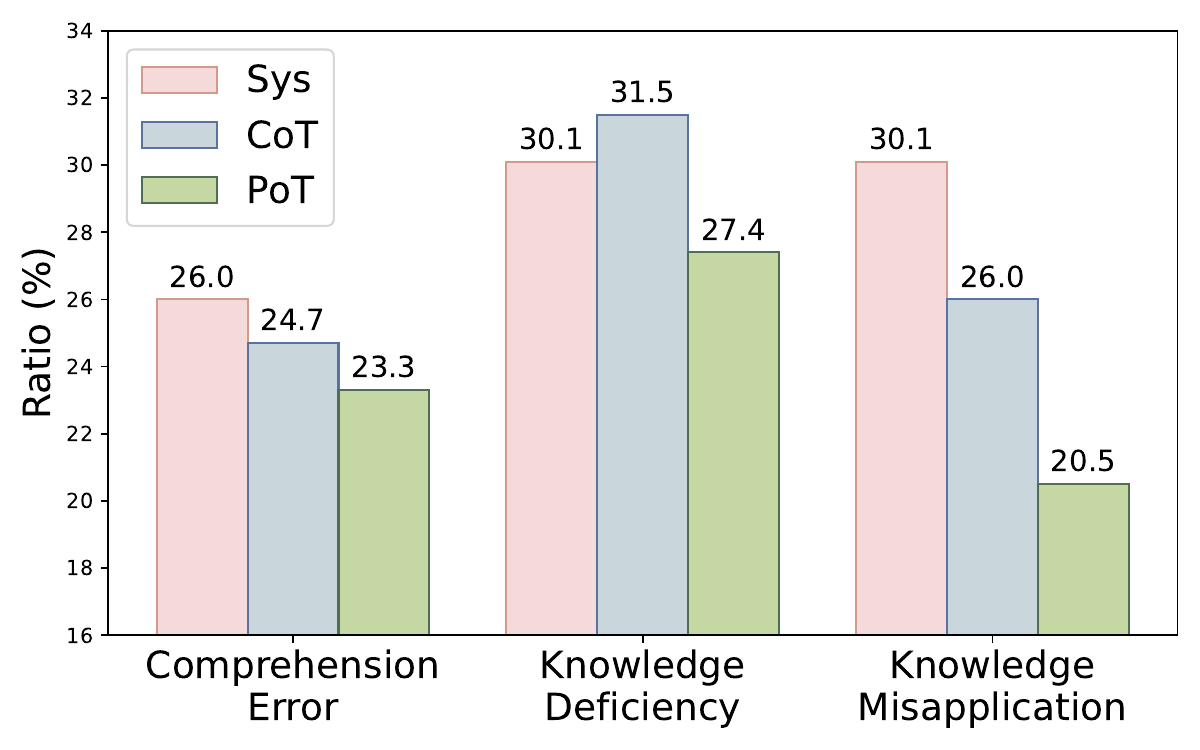}
  \caption{Distribution of the proposed three error types across Sys, CoT, and PoT baselines.
  }
  \vspace{-2mm}
  \label{fig: error ana baselines}
\end{figure}

\begin{figure}[t!]
  \centering
  \includegraphics[width=\columnwidth]{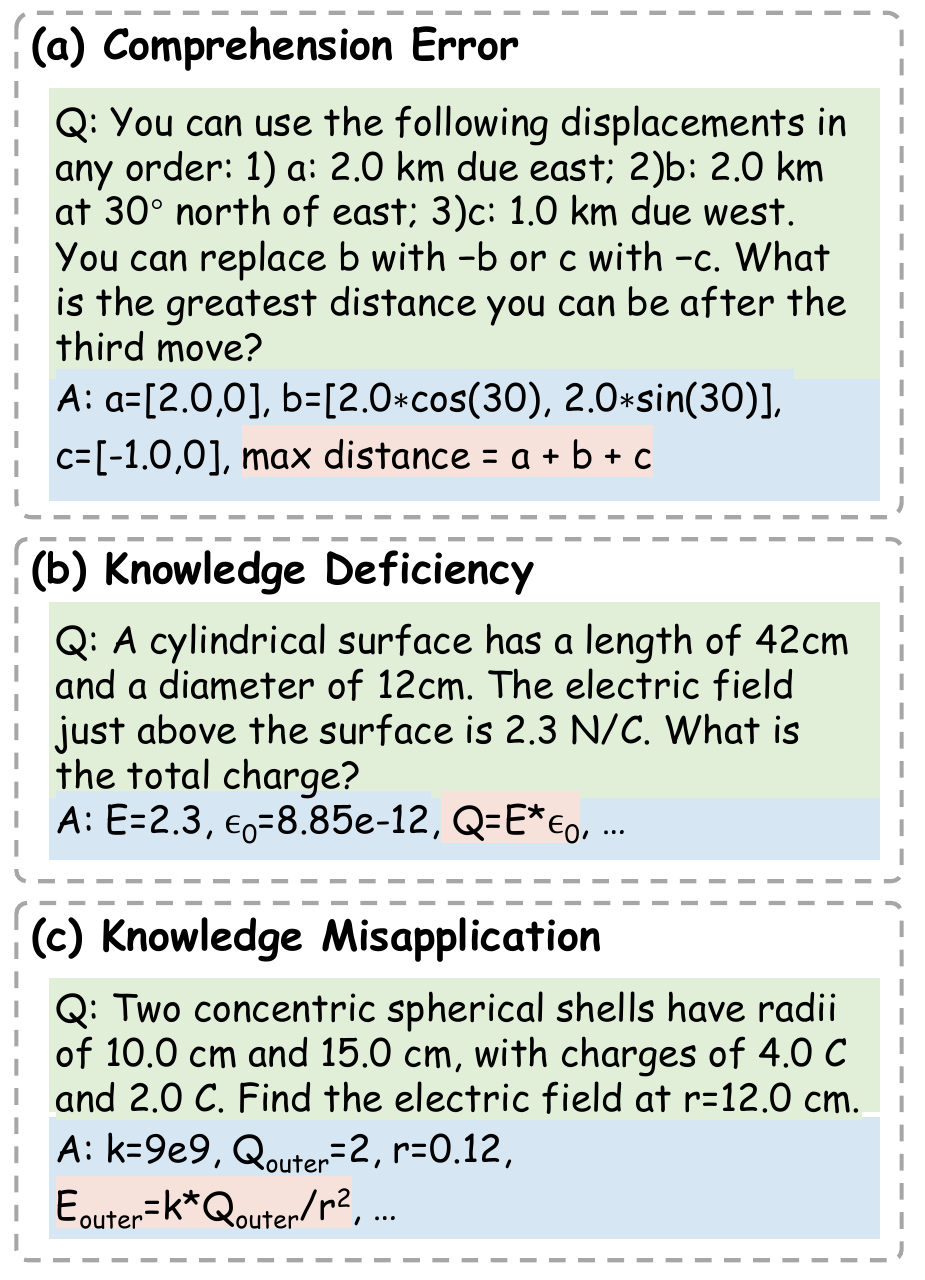}
  \caption{Example for each error type, where the red highlighted parts indicate errors.\\
  }
  \vspace{-8mm}
  \label{fig: error examples}
\end{figure}

LLMs often encounter difficulties and make errors when tackling complex physics problems. To investigate the primary factors behind these errors, we manually analyze numerous error cases in the fund dataset from the SciBench benchmark using three representative baselines, \textit{i.e.} System (Sys), Chain of Thought (CoT), and Program of Thought (PoT)~\cite{DBLP:journals/corr/abs-2211-12588}. The detailed introduction of these baselines can be found in Sec.~\ref{sec: baselines}. Following the paradigm in SciBench, experiments are conducted using GPT-3.5-turbo~\cite{GPT-3.5}. Two undergraduate students with strong backgrounds in physics are enlisted to categorize these errors into three types as follows:
\vspace{-1mm}
\begin{itemize}[leftmargin=*]
    \item \textbf{Comprehension Error} refers to misunderstanding of the problem, including misinterpreting the context and identifying incorrect assumptions.
    \vspace{-1mm}
    \item \textbf{Knowledge Deficiency} indicates a lack of the necessary knowledge required to solve the problems effectively and accurately.
    \vspace{-1mm}
    \item \textbf{Knowledge Misapplication} involves the incorrect use of relevant knowledge, such as misinterpretation of key concepts, incorrect formula translation, and other related errors.
\end{itemize}
\vspace{-1mm}
Each error type is illustrated with an example in Fig.~\ref{fig: error examples}.
To further explore the frequency of these error types, we compile statistics on error cases on the fund dataset across three baselines. 
The error distribution for each error type is shown in Fig.~\ref{fig: error ana baselines}, suggesting the two main reasons for LLMs' recurring shortcomings: 
(i) insufficient physics knowledge in LLMs;
(ii) challenges in correctly applying the knowledge to solve physics problems.
Motivated by the observation, we propose \Ours{} to mitigate these issues, detailed in Sec.~\ref{sec: approach}.

\section{\Ours{}}
\label{sec: approach}

\begin{figure*}[t!]
  \centering
  \includegraphics[width=\linewidth]{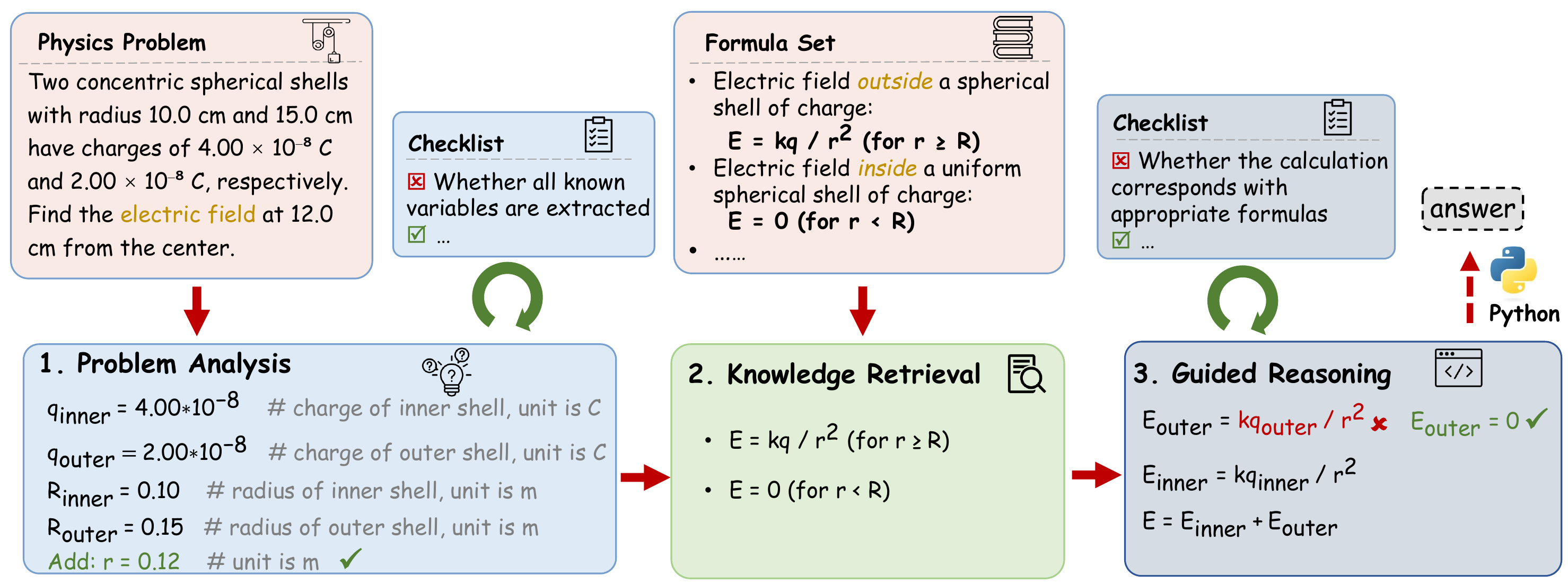}
  \caption{Illustration of \Ours{}, solving a physics problem using LLMs with the help of the formula set and checklists. The approach contains three stages: problem analysis, knowledge retrieval, and guided reasoning.
  }
  \vspace{-2mm}
  \label{fig: method}
\end{figure*}

Physics problem reasoning involves solving problems with domain knowledge and necessary steps. The inputs include a problem $P$ and a constructed formula set $F$. The desired output is the calculated result $R$ in the form of the target unit.


We introduce details of the workflow of our approach, construction of the formula set, and design of checklists in Sec.~\ref{sec: workflow},~\ref{sec: formula set} and~\ref{sec: checklist} respectively.

\subsection{Workflow}
\label{sec: workflow}

\Ours{} is divided into 3 stages: problem analysis, formula retrieval, and guided reasoning.

As shown in Fig.~\ref{fig: method}, given a physics problem $P$, along with our formula set $F$, \Ours{} first comprehends $P$ and then extracts known variables $V=\{v_0, v_1, ..., v_n\}$ from it as incomplete Python code. Here, each variable $v_i$ is roughly defined in a single line of code, followed by comments explaining the specific meaning of the variable and its corresponding unit. After this initial extraction, a checklist $CL_{PA}$ is used to review and refine the analysis process. The details of this checklist are introduced in Sec.~\ref{sec: checklist}.

The next step is formula retrieval, which is conducted hierarchically. Formulae are categorized into various fields of physics. Each formula is accompanied by a brief description, formula content, and definitions of involved variables. Based on problem text $P$, the LLM first determines relevant fields and retrieves formulae from them. It then selects formulae $F=\{f_0, f_1, ..., f_m\}$ related to $P$ and writes them as Python code comments, supplementing the extracted variables $V$.

Subsequently, \Ours{} completes the Python code $C$ with precise reasoning using extracted variables $V$ and formulae $F$, printing the target variable in the required unit at the end. Following this, another checklist $CL_{GR}$ is utilized for guiding and refining the reasoning process. The refined code $C'$ is then executed, and the output variable is taken as the predicted answer.

\vspace{-2mm}
\subsection{Formula Set Construction}
\label{sec: formula set}
This section outlines the principles and process of constructing our formula set, aiming to provide explicit knowledge of physics formulae to enhance knowledge acquisition in LLMs.

\subsubsection{Collection Principles}
\noindent \textbf{Covering common physics problems. }
We expect the formula set to cover common college-level physics problems. To achieve this, we select three prevalent college physics textbooks as collection sources and identify four representative fields.

\noindent \textbf{Clarifying detailed introduction and variable definition. }
Another principle is to explicitly explain the correct conditions for formula applications and provide clear definitions of involved variables. 
Even with the formula known, correct reasoning is impossible without applying it in an appropriate context or accurately substituting the variables. To enhance knowledge acquisition, each formula in our set is accompanied by a detailed introduction and precise variable definitions.

\subsubsection{Formula Collection}
We choose three physics textbooks to collect formulae following SciBench: \textit{Fundamentals of Physics}~\cite{halliday2013fundamentals}, \textit{Statistical Thermodynamics}~\cite{engel2019thermodynamics}, and \textit{Classical Dynamics of Particles and Systems}~\cite{greiner2003classical}. These textbooks are representative of the field of physics, containing abundant knowledge of formulae.

\begin{figure}[t]
  \centering
  \includegraphics[width=0.8\linewidth]{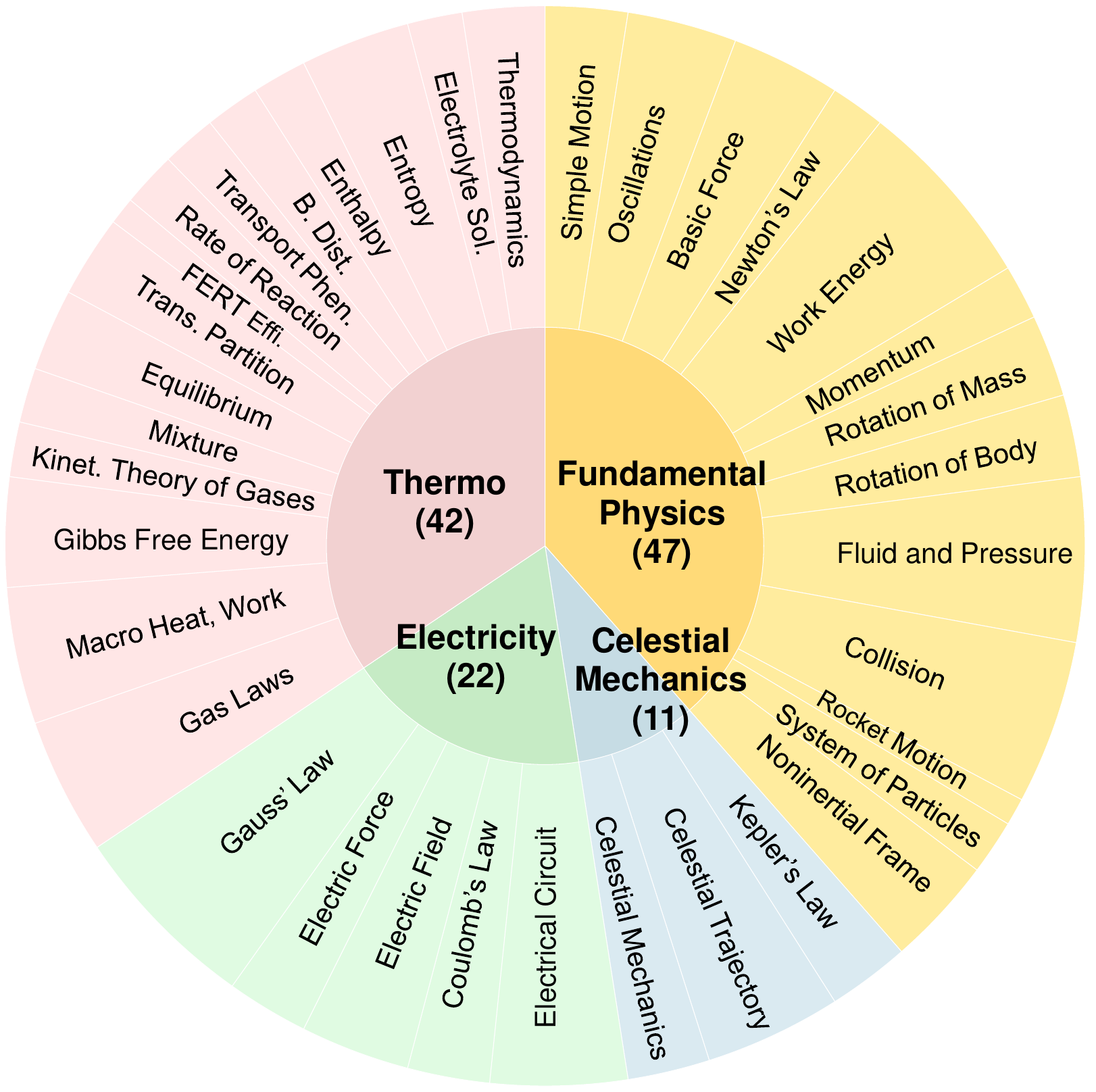}
  \caption{Subfields in our formula set.
  }
  \label{fig: formula dist}
\end{figure}

\definecolor{customcolbacktitle}{HTML}{DEEBF7}
\definecolor{customcolframe}{HTML}{112D4E}
\definecolor{customcolback}{HTML}{D5DCE4}
\definecolor{highlight}{HTML}{1679AB}
\begin{figure}[t]
    \small
    \centering
    \begin{tcolorbox}[colback=gray!10,colframe=customcolframe,
        left=7pt,right=7pt,top=5pt,bottom=5pt,
                title=\textbf{Formula Example},
        colbacktitle=customcolbacktitle,
        coltitle=black]
        
        \raggedright
        \textcolor{highlight}{\textbf{name}}: "Kepler's Third Law"\\
        \textcolor{highlight}{\textbf{content}}: "T ** 2 = (4 * pi ** 2 / G * M) * a ** 3",\\
        \textcolor{highlight}{\textbf{variables}}:\\
        "T": "Period of the orbit",\\
        "G": "Gravitational constant",\\
        "M": "Mass of the central body",\\
        "a": "Semi-major axis of the orbit"\\
    \end{tcolorbox}
    
    \caption{An example of our formula annotation.}
    \label{fig: formula example}
    \vspace{-2mm}
\end{figure}
Initially, we collect $1124$ formulae from these textbooks and refine the collection by removing: (i) repetitive formulae, such as those differing only in the order of multiplication and division; (ii) intermediate calculation processes that do not constitute scientific formulae; and (iii) conclusions from individual example problems that are not universal. After the refinement, we identify $122$ representative formulae. 
As shown in Fig.~\ref{fig: formula dist}, we categorize these formulae into 36 subfields, which are further classified into four major fields: Fundamental Physics, Celestial Mechanics, Electricity, and Thermodynamics. 
For each formula, we include a detailed introduction and definitions of the involved variables to enhance knowledge application. Fig.~\ref{fig: formula example} provides an example of a formula annotation. 

\newcommand{\cmark}{\textcolor{green}{\checkmark}} 
\newcommand{\xmark}{\textcolor{red}{\ding{55}}}     

\begin{table}[t]
\small
\centering
\resizebox{\linewidth}{!}{
\begin{tabular}{@{}lccc@{}}
\toprule
\textbf{Formula Set}     & \textbf{Physics} & \textbf{English} & \textbf{Theorem} \\
\midrule
Math23K-F~\cite{DBLP:conf/kdd/LiuHMLCSL23}    & \xmark    & \cmark    & \xmark        \\
MAWPS-F~\cite{DBLP:conf/kdd/LiuHMLCSL23}  &    \xmark     &  \cmark    &        \xmark             \\
FormulaQA~\cite{DBLP:journals/corr/abs-2402-12692}     &    \cmark     &     \xmark    &          \xmark           \\
Ours        &   \cmark      &      \cmark   &         \cmark           \\
\bottomrule
\end{tabular}
}

\caption{Comparison with other formula sets.}
\label{tab: formulaset_comparison}
\end{table}

\subsubsection{Comparison with Existing Dataset}
To the best of our knowledge, there are few datasets of physics formulae designed for complex physics problem reasoning. 
Previous formula-based datasets mainly consist of commonsense formulae. For instance, in the MAWPS-F~\cite{DBLP:conf/kdd/LiuHMLCSL23} dataset, the most frequently used formula is $total\,amount = unit\,amount * total\,number$, which is basic commonsense knowledge. 

The most relevant work to ours is FormulaQA~\cite{DBLP:journals/corr/abs-2402-12692}, a question-answering numerical reasoning dataset. It contains annotated problems selected from Chinese junior high school physics examinations. However, the formulae in it are primarily abstract expressions derived from individual example problems rather than scientific theorems in physics, such as $Degree\,of\,temperature\,increase = Final\,temperature - Initial\,temperature$. This makes them heavily dependent on the problem and limits their potential for extensibility. 

A detailed comparison between our formula set and other existing datasets is presented in Table~\ref{tab: formulaset_comparison}.

\subsection{Checklist}
\label{sec: checklist}
This section outlines the principles and procedures for designing effective checklists, which include detailed instructions for identifying and correcting prone errors. These checklists are utilized to help LLMs self-improve by enabling them to apply physics knowledge more accurately.

\subsubsection{Design Principles}
\textbf{Focusing on common mistakes of physics problems. }
We propose checklists to help avoid common mistakes LLMs often make. By thoroughly analyzing numerous incorrect processes in physics problems, we identify and summarize representative errors to include in the checklists. For example, LLMs often confuse vectors with scalars, leading to incorrect reasoning. The checklist includes steps to verify the text and distinguish vectors from scalars.

\noindent
\textbf{Enhancing LLMs' application concisely. }
Our goal is using checklists to enhance LLMs' accurate application of domain knowledge. Merely acquiring the necessary physics knowledge is not enough to successfully solve a complex problem. Additionally, we found that providing too many instructions can distract LLMs and hinder their ability to apply knowledge effectively. Therefore, our checklists are designed to be both instructive and concise, offering clear guidance without leading to distraction.

\subsubsection{Design Procedure}
Since the process of solving physics problems is sequential, an error in an earlier step may lead to mistakes in subsequent reasoning.
We observe that errors primarily occur during problem comprehension and knowledge application.
Motivated by this, we propose separate checklists for both problem comprehension and calculation stages, trying to ensure the accuracy of each step.

We design checklists based on the errors observed. By testing several LLM-based reasoning methods on abundant physics problems, we identify key points that need to be carefully checked and include them in the checklists. 
Additionally, common errors vary across different fields. For instance, in electronics problems, LLMs often confuse Coulomb's constant $k$ with permittivity of free space $\epsilon_0$ when calculating electronic fields. To address such issues, we design specialized checklists for various fields. 
Namely, our checklist examples are shown in Fig.~\ref{fig: checklist variables} and Fig.~\ref{fig: checklist code}.
\begin{figure}[t]
    \small
    \centering
    \begin{tcolorbox}[colback=gray!10,colframe=customcolframe,
        left=7pt,right=7pt,top=5pt,bottom=5pt,
        title=\textbf{Checklist for Problem Analysis},
        colbacktitle=customcolbacktitle,
        coltitle=black]
        
        You are an excellent physics teacher. Given a physics problem and its extracted known variables, please examine the variables using the following checklist and revise it:\\
    \\
    \textcolor{highlight}{**Checklist**:}\\
    - Understand the problem, whether all known variables are extracted.\\
    - Whether the variable names are clear and can distinguish the different objects to which they may belong.\\
    - Whether the variables have been converted to SI units.\\
    - Check if the variables have confused vectors with scalars.\\
    \\
    Return the correct known variables in the format of Python code, encasing it within triple back-ticks for clarity.\\
    
    \{Problem text\}\\
    \{Initial extracted variables\}
    \end{tcolorbox}

\caption{Checklist for problem comprehension, verifying extracted variables and units}
\label{fig: checklist variables}
\vspace{-2mm}
\end{figure}
\begin{figure}[t]
    \small
    \centering
    \begin{tcolorbox}[colback=gray!10,colframe=customcolframe,
        left=7pt,right=7pt,top=5pt,bottom=5pt,
        title=\textbf{Checklist for Guided Reasoning},
        colbacktitle=customcolbacktitle,
        coltitle=black]

You are an excellent physics teacher. Given a physics problem and its Python code, please review the code using the following checklist and improve it:\\
\\
\textcolor{highlight}{**Checklist**:}\\
- If the problem text is correctly understood and solved.\\
- Whether the calculation corresponds with correct, appropriate formulas.\\
- Whether the variables and constants are correctly defined before use, ensuring that constants are not confused.\\
- If the problem is solved, whether the code prints the first target variable at the end.\\
- Check the unit, whether the target variable is printed in the required unit.\\
\\
Please reiterate your code and return the improved code. Encasing the code within triple back-ticks for clarity.\\

\{Problem text\}\\
\{Initial solving process\}
    \end{tcolorbox}

\caption{Checklist for calculation process, verifying physics knowledge application.}
\label{fig: checklist code}
\vspace{-2mm}
\end{figure}

\definecolor{mycolor1}{RGB}{247,226,219}
\definecolor{mycolor2}{RGB}{142,141,200}
\definecolor{mycolor3}{RGB}{71,133,90}
\definecolor{mycolor4}{RGB}{255,255,255}

\newcolumntype{X}{>{\columncolor{mycolor1}}c}
\newcolumntype{Y}{>{\columncolor{mycolor2!50}}c}
\newcolumntype{Z}{>{\columncolor{mycolor3!50}}c}
\newcolumntype{K}{>{\columncolor{mycolor4}}c}

\begin{table*}[t!]
\small
\centering
\resizebox{\linewidth}{!}{
\begin{tabular}{l|XXXYXXXYXXXY}
\toprule
\multirow{2}{*}{\textbf{Method}} & \multicolumn{4}{c}{\textbf{GPT-3.5-turbo}} & \multicolumn{4}{c}{\textbf{Llama3-70B}}   & \multicolumn{4}{c}{\textbf{GPT-4-turbo}} \\ 
                        & \multicolumn{1}{c}{\textbf{fund}}   & \multicolumn{1}{c}{\textbf{thermo}}  & \multicolumn{1}{c}{\textbf{class}}  & \multicolumn{1}{c}{\textbf{Avg.}}  & \multicolumn{1}{c}{\textbf{fund}}   & \multicolumn{1}{c}{\textbf{thermo}}  & \multicolumn{1}{c}{\textbf{class}}  & \multicolumn{1}{c}{\textbf{Avg.}} & \multicolumn{1}{c}{\textbf{fund}}   & \multicolumn{1}{c}{\textbf{thermo}}  & \multicolumn{1}{c}{\textbf{class}}  & \multicolumn{1}{c}{\textbf{Avg.}}\\
\midrule
System      & 13.7   & 11.9    & 2.1   &   9.3    & 5.5  & 14.9   & 14.9  &  11.8    & 42.5  &  41.8  & 25.5   &   36.6  \\
CoT   & 17.8  & 9.0     & 4.3    &   10.3    & 27.4 & 16.4   & \underline{19.2}  &   21.0   &   50.7    &    26.9  & 25.5   &  34.4 \\
PoT    & 28.8   & 13.4     & 12.8    &   18.3    & \underline{45.2} & \textbf{34.3}   & 14.9  &    \underline{31.5}  & \underline{63.0}  &   \underline{46.3}  & \underline{31.9}   & \underline{47.1}  \\
PoT + Self-Correction   & \underline{30.1}   & 11.9    & 10.6   &    17.6   & 34.3 & 20.9   & 14.9  &    23.4  &   24.7    &    25.4     &     19.2   &   23.1   \\
PoT + Self-Refine    & \underline{30.1}   &     \underline{19.4}    &  \underline{14.9}      &      \underline{21.5} & 30.1 &    26.9    & 17.0 &    24.7  &  53.4 &   23.9  &  \underline{31.9}  &  36.4 \\ \midrule
\Ours{} (Ours)   & \textbf{38.4}   &    \textbf{25.4}     & \textbf{23.4}   &    \textbf{29.1}   & \textbf{50.7} &   \underline{29.9}     & \textbf{25.5}  &  \textbf{35.4}   & \textbf{70.0}  &   \textbf{50.8}      & \textbf{38.3}   &   \textbf{53.0}  \\
\bottomrule
\end{tabular}
}
\caption{Accuracy results (\%) on three SciBench physics datasets: fund, thermo, and class under few-shot setting. \textbf{Boldface} numbers highlight the best results; \underline{underlined} numbers represent the second-best.}
\label{tab: main}
\end{table*}

\vspace{-1mm}
\section{Experiments}
\label{sec: experiments}
\subsection{Dateset}
We conduct experiments on three physics datasets from the SciBench benchmark, systematically examining the reasoning capabilities required to solve complex problems. These datasets cover a wide range of fields in physics, including motion, kinetic energy, electronics, and so on. The diversity enhances the comprehensiveness of evaluating LLMs' abilities in physics problem reasoning.

The datasets are manually collected from various college-level physics textbooks and are selected for challenging, free-response answers. Each dataset is divided into two parts, $P_s$ and $P_w$. $P_w$ contains most of the problems without solutions, while $P_s$ comprises several problems with detailed solutions, which can be used as examples in a few-shot setting. Detailed statistics are presented in Table~\ref{tab: dataset statistics}.
\begin{table}[t!]
\small
\centering

\begin{tabular}{lccc}

\toprule
\textbf{Dataset} & \textbf{Field }         & \textbf{\# P} & \textbf{\# S} \\
\midrule
fund    & electronics        & 83          & 10           \\
thermo  & thermodynamics     & 84          & 17           \\
class   & classical dynamics & 54          & 7             \\
\bottomrule
\end{tabular}
\caption{Dataset Statistics. $\# P$ and $\# S$ represent the number of total problems and problems with detailed solutions, respectively.}
\label{tab: dataset statistics}
\vspace{-4mm}
\end{table}

\subsection{Implementation Details}

We test on both open-source and close-source LLMs, including GPT-4~\cite{DBLP:journals/corr/abs-2303-08774}, GPT-3.5~\cite{GPT-3.5}, and Llama 3~\cite{llama3}.
For evaluation metrics, we follow~\citet{DBLP:journals/corr/abs-2307-10635} to compare the predicted answers with the correct answers, allowing a relative tolerance of 5\%.
Few-shot examples are randomly selected within each textbook, ensuring consistency with SciBench.
The temperature is set to zero to reduce randomness.
Llama3 is executed on an NVIDIA A800-SXM4-80GB GPU, with a time cost of approximately two hours for these datasets. The other two models are accessed through OpenAI API.

\vspace{-1mm}
\subsection{Baselines}
\label{sec: baselines}
We evaluate multiple baselines under the few-shot setting, following the SciBench evaluation paradigm to ensure a rigorous evaluation process:
\vspace{-2mm}
\begin{itemize}[leftmargin=*]
    \item \textbf{System} refers to directly feeding the problem with instructions describing question types.
    \vspace{-3mm}
    \item \textbf{CoT} points to the Chain of Thought strategy, where the model outputs the reasoning process step by step before providing a predicted answer.
    \vspace{-7mm}
    \item \textbf{PoT} guides LLMs to write and then execute a Python program to determine the final answer. 
    \vspace{-3mm}
    \item \textbf{PoT + Self-Correction}~\citet{DBLP:journals/corr/abs-2310-01798} asks LLMs to rectify initial PoT responses.
    \vspace{-3mm}
    \item \textbf{PoT + Self-Refine} requests the model to provide feedback on its initial output by PoT and uses the feedback to refine itself ~\cite{DBLP:conf/nips/MadaanTGHGW0DPY23}. The iteration number is set to 1 in this case.
\end{itemize}

\section{Result Analysis}
\vspace{-2mm}
\subsection{Reasoning Quality}
Table~\ref{tab: main} presents the accuracy of different models on the test set of the three datasets. Based on the results, we have the following observations: 

\textbf{\Ours{} achieves superior accuracy. }
As shown in the table, \Ours{} outperforms all baselines, improving average accuracy from 21.5\% to 29.1\% on GPT-3.5-turbo, from 31.5\% to 35.4\% on Llama3:70b, and from 47.1\% to 53.0\% on GPT-4-turbo. 
It is worth noting that \Ours{} can be regarded as an enhanced form of PoT. The improvements indicate that \Ours{} benefits from the formula set to provide physics knowledge and from the checklists, which enhance knowledge application.

\textbf{\Ours{} gains considerable advancement in complex problems. }
We observe that \Ours{} performs better on fund dataset compared to the other datasets. One possible explanation is that fund contains problems in more fundamental fields, such as gravitation, which is more frequently encountered in LLMs' training data, leading to potentially better performance. Also, as shown in Table~\ref{tab: formula usage}, a larger proportion of problems in fund requires formulae usage and the average number of formulae per problem is also higher. These differences highlight \Ours{}'s capability in complex physics problem reasoning.

\begin{figure}[t]
  \centering
  \includegraphics[width=\columnwidth]{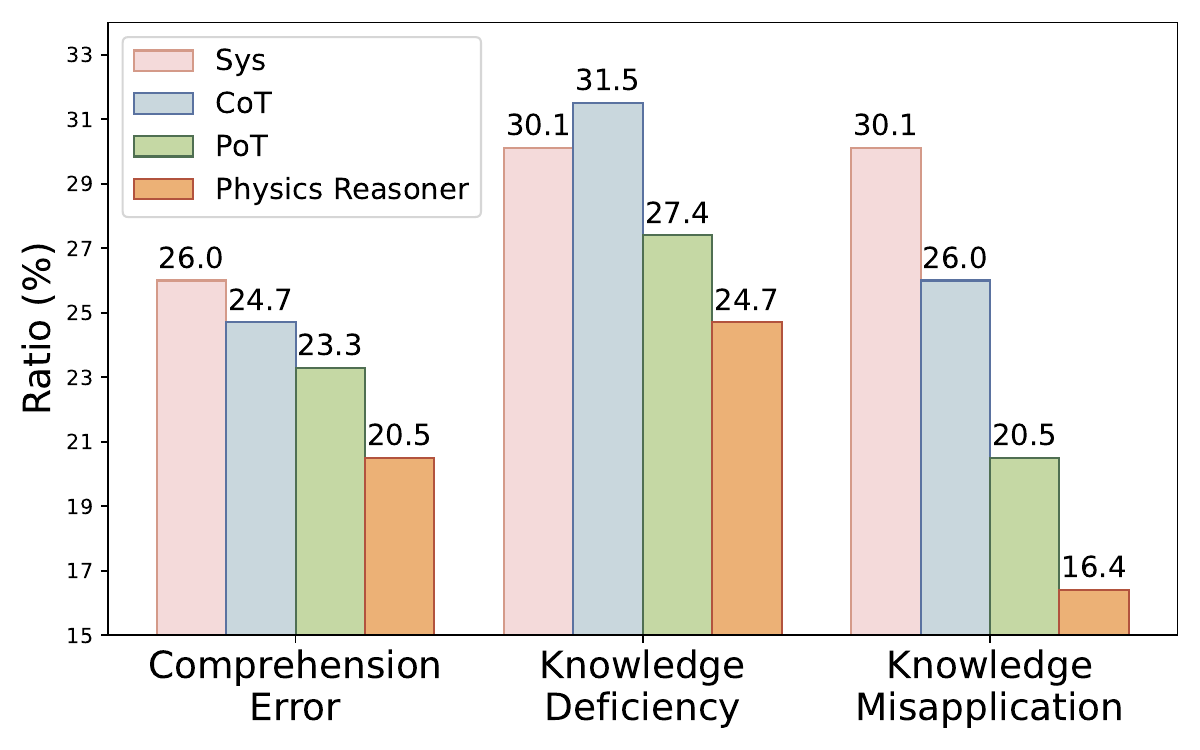}
  \caption{Proportion of each error type for Physics Reasoner and other three baselines.
  }
  \label{fig: error ana ours}
  \vspace{-2mm}
\end{figure}

We further conduct manual analysis of error cases for \Ours{} on the fund dataset with the GPT-3.5-turbo model following the error types defined in Sec.~\ref{sec: limitation}. The results are shown in Figure~\ref{fig: error ana ours}, revealing that \Ours{} significantly reduces all three types of errors.

\vspace{-6mm}
\revise{
\subsection{Tool-Argumented Baselines}
For a more thorough comparative analysis, we also conducted supplementary experiments with two representative tool-argumented baselines, \textit{i.e.} CREATOR~\cite{qian2023creator} and Chameleon~\cite{DBLP:conf/nips/LuPCGCWZG23}, utilizing GPT-3.5-turbo. The parameters are aligned with the source code separately, and both the few-shot examples and metrics are consistent with our main experiments. The accuracy results (\%) are represented in Table~\ref{tab: tool argumented baselines}.
\begin{table}[t]
\small
\centering
\resizebox{\linewidth}{!}{
\begin{tabular}{lcccc}
\toprule
\textbf{Method}     & \textbf{fund} & \textbf{thermo} & \textbf{class} & \textbf{Avg.} \\
\midrule
CREATOR                 & 21.9 & 17.9   & 8.5   & 16.1 \\
Chameleon               & 32.9 & 14.9   & 19.2  & 22.3 \\
\midrule
\textbf{Physics Reasoner (Ours)} & \textbf{38.4} & \textbf{25.4}   & \textbf{23.4}  & \textbf{29.1} \\
\bottomrule
\end{tabular}
}
\caption{Results (\%) of tool-argumented baselines.}
\label{tab: tool argumented baselines}
\vspace{-4mm}
\end{table}\\
Physics Reasoner outperforms other baselines across all datasets, achieving an average improvement of 6.73\%. The consistent advantage underscores the significant efficacy of our approach.
}

\vspace{-2mm}
\revise{
\subsection{Token Cost Comparison}
We count the token cost of different methods with GPT-3.5-turbo in Table~\ref{tab: token cost}. The total tokens of Physics Reasoner are fewer than the previous tool-argument methods.  
It is worth noting that our approach is not iterative and requires a fixed 4-step process per problem. In contrast, Chameleon requires 5-6 steps and CREATOR demands 4-5 steps, leading to a higher overall token expenditure.
\newcolumntype{L}[1]{>{\raggedright\let\newline\\\arraybackslash\hspace{0pt}}m{#1}}
\newcolumntype{C}[1]{>{\centering\let\newline\\\arraybackslash\hspace{0pt}}m{#1}}

\begin{table}[t]
\small
\centering
\resizebox{\linewidth}{!}{
\begin{tabular}{lcccc}
\toprule
\textbf{Method}     & \textbf{fund} & \textbf{thermo} & \textbf{class} & \textbf{Avg.} \\
\midrule
CREATOR                 & 571  & 533    & 400   & 502  \\
Chameleon               & 538  & 520    & 289   & 449  \\
\midrule
Physics Reasoner (Ours) & \textbf{470}  & \textbf{435}    & \textbf{253}   & \textbf{386} \\
\bottomrule
\end{tabular}
}
\caption{Token cost of tool-argumented baselines (in thousands of tokens, k tokens).}
\label{tab: token cost}
\vspace{-2mm}
\end{table}
}

\subsection{Ablation Study}

\newcolumntype{C}[1]{>{\centering\arraybackslash}p{#1}}

\vspace{2mm}
\begin{table}[t]
\small
\centering
\begin{tabular}{@{}C{1.8cm}C{1.25cm}C{1.25cm}|C{1.5cm}@{}}
\toprule
\textbf{Formula Set} & \textbf{CL\_VE}   & \textbf{CL\_SP} & \textbf{Accuracy} \\
\midrule
\cmark     &     \cmark       &       \cmark        &    38.4\\
\midrule
\cmark      &     \cmark         &     \xmark  &   35.2  \\
 \cmark       &    \xmark    &     \xmark     &   30.1  \\
 \xmark        &    \xmark  &    \xmark   &   28.8  \\ 
     
\bottomrule
\end{tabular}
\caption{Ablation study results (\%) of \Ours{} on fund dataset with GPT-3.5-turbo, in which CL\_VE represents checklist for variable extraction, CL\_SP represents checklist for solving process.}
\label{tab: ablation}
\end{table}
\vspace{-1mm}
We further investigate the effectiveness of each component in \Ours{}, focusing on three key components: the formula set, the checklist for problem comprehension, and the checklist for calculation, which are sequentially removed.

Table~\ref{tab: ablation} presents the results of the ablation study conducted with GPT-3.5-turbo on the fund dataset, offering the greatest diversity of questions. We observe that the inclusion of each component leads to a consistent improvement, underscoring the necessity of each component in our method.

\subsection{Necessity of Formulae in Problem Solving}
\newcolumntype{C}[1]{>{\centering\arraybackslash}p{#1}}

\begin{table}[t]
\small
\centering
\begin{tabular}{@{}l|C{1.8cm}C{1.6cm}C{1.6cm}@{}}
\toprule
\textbf{Dataest} & \textbf{Num. Form.} & \textbf{\% Form.} & \textbf{Avg. Num.} \\
\midrule
fund    & 58                      & 79.45                       & 1.58                      \\
thermo  & 33                      & 49.25                       & 0.63                      \\
class   & 35                      & 74.47                       & 0.89                     \\
\bottomrule
\end{tabular}
\caption{Formula usage across datasets: "Num. Form." for the number of problems with formulae, "\% Form." for the percentage of problems with formulae, and "Avg. Num." for the average number of formulae per problem.}
\label{tab: formula usage}
\end{table}

From the analysis of incorrect examples in Sec.~\ref{sec: limitation}, we observe that LLMs often make mistakes due to a lack of relevant knowledge. After the construction of our formula set, we are interested in whether the formulae are utilized in the reasoning process. Formula usage is summarized in Table~\ref{tab: formula usage}, from which we observe the following key points:

\textbf{Most problems necessitate formulae. }
The vast majority of problems in the fund and class datasets rely on formulae from our formula set, as do approximately half of the problems in the thermo dataset. Overall, most problems across the three datasets require the use of physics formulae from our formula set for solutions.

\textbf{The number of formulae required for each problem varies. }
While some simple problems can be solved with a single formula, others may require up to four. Generally, problems with fewer formulae are easier to solve. Complex problems require a thorough application of multiple formulae and detailed analysis, increasing their difficulty.

\subsection{Checklists Enhance Effective Knowledge Application}
\newcolumntype{L}[1]{>{\raggedright\arraybackslash}p{#1}}

\begin{table}[t]
\small
\centering
\resizebox{\linewidth}{!}{
\begin{tabular}{@{}lcc@{}}
\toprule
\textbf{Method }         & \textbf{Wrong2Correct\textuparrow }& \textbf{Correct2Wrong\textdownarrow} \\
\midrule
PoT+Self-Correction &   0.0  &   6.3  \\
PoT+Self-Refine     &  8.7   &   26.0  \\ \midrule
\Ours{} (Ours)        &  \textbf{16.7}   &    \textbf{3.9}    \\
\bottomrule
\end{tabular}
}
\caption{Ratio (\%) to correct incorrect problems (Wrong2Correct) and mislead answers (Correct2Wrong) of different methods on fund dataset with GPT-4-turbo.}
\label{tab: checklist analysis}
\vspace{-4mm}
\end{table}

To assess the efficacy of checklists in knowledge application, we carefully examine the results of \Ours{} and two self-improvement methods, \textit{i.e.} self-correction and self-refine on the fund dataset using the GPT-4-turbo model. The proportion of initial errors that are corrected and misled problems are shown in Table~\ref{tab: checklist analysis}.

Our method corrects 16.67\% of incorrect reasoning processes, whereas the other two baselines correct only 8.7\% and 0.0\%. Additionally, the number of misled problems notably decreases with our method, indicating that \Ours{} significantly enhances knowledge application.



\section{Conclusion}

In this paper, we propose \Ours{}, a novel framework for physics problem reasoning that integrates knowledge acquisition through a formula set and knowledge application with meticulously designed checklists. We compile a representative formula set and design detailed checklists to address issues related to the lack of knowledge or incorrect knowledge application. This approach significantly improves accuracy in solving physics problems and achieves state-of-the-art performance. Given the limited research on physics problem reasoning, we advocate for further studies to explore the potential of this field, aiming to advance the effectiveness of LLMs in complex problem-solving scenarios.

\section*{Limitations}
In this paper, we propose \Ours{}, a novel framework to mitigate knowledge gaps and knowledge misapplication in physics problem reasoning. Despite the promising performance of our approach, we acknowledge that there are areas for improvement and opportunities for future research.

Firstly, we conduct experiments on a limited number of datasets from SciBench using three models. Our results may not fully reflect the reasoning abilities of LLMs under different conditions. Future research should examine a broad range of physics problem datasets and model types to provide a more comprehensive analysis.

Secondly, our formula set covers a limited range of physics fields due to the manual effort required for its construction. Expanding our formula set to include more physics fields would be beneficial and could provide deeper insights to further study.

\section*{Aknowledgements}
We gratefully acknowledge the anonymous reviewers for their thoughtful feedback.
This work was also supported by the National Science and Technology Major Project (No. 2022ZD0114903), the Natural Science Fundation of China (NSFC. No. 62476149), and the Guoqiang Institute of Tsinghua University, with Grant No. 2020GQG0005.

\section*{Ethical Considerations}
This article adheres to the ACL Code of Ethics. According to our knowledge, our work constitutes foundational research, and we do not identify any significant risks related to malicious harm, environmental impact, fairness, or privacy concerns.

\bibliography{custom}

\clearpage
\appendix
\label{sec:appendix}
\section{Details for Formula Set Construction}
\subsection{Collection Details}
\noindent
\textbf{Classification: }
We chose three textbooks as the sources for our formula collection to guarantee correctness and comprehensiveness. Two undergraduate students with strong backgrounds in physics summarize these textbooks and re-categorize them into four representative fields, \textit{i.e.} fundamental physics, celestial mechanics, electricity, and thermodynamics (thermo). Furthermore, each field is classified into several subfields, which are more refined and specific topics.

\noindent
\textbf{Selection: }
We focus on scientific theorems, removing problem-specific formulae and formulae in different formats with the same meanings. 

\noindent
\textbf{Annotations: }
Each formula is annotated in three parts: a brief description of the formula, formula content, and clear definitions of involved variables.

\subsection{Detailed Statistics}
The detailed statistics of our formula set, including the distribution and categorization of formulae across various physics fields and subfields, are shown in Table~\ref{tab: appendix formula dist}.
\begin{table}[t]
\small
\centering
\begin{tabular}{lcc}
\toprule
Field                                                                           & Subfield                 & Number \\
\midrule
\multirow{13}{*}{\begin{tabular}[c]{@{}l@{}}Fundamental \\ Physics\end{tabular}} & Simple Motion           & 3   \\
                                                                                 & Oscillations            & 4   \\
                                                                                 & Basic Force             & 4   \\
                                                                                 & Newton's Law            & 2   \\
                                                                                 & Work Energy             & 7   \\
                                                                                 & Momentum                & 2   \\
                                                                                 & Rotation of Mass        & 3   \\
                                                                                 & Rotation of Body        & 3   \\
                                                                                 & Fluid and Pressure      & 6   \\
                                                                                 & Collision               & 6   \\
                                                                                 & Rocket Motion           & 1   \\
                                                                                 & System of Particles     & 2   \\
                                                                                 & Noninertial Frame       & 4   \\
\midrule
\multirow{3}{*}{\begin{tabular}[c]{@{}l@{}}Celestial \\ Mechanics\end{tabular}}  & Kepler's Law            & 3   \\
                                                                                 & Celestial Trajectory    & 5   \\
                                                                                 & Celestial Mechanics     & 3   \\
\midrule
\multirow{5}{*}{Electricity}                                                     & Electrical Circuit      & 5   \\
                                                                                 & Coulomb's Law             & 3   \\
                                                                                 & Electric Field          & 4   \\
                                                                                 & Electric Force, Energy  & 3   \\
                                                                                 & Gauss' Law               & 7   \\
\midrule
\multirow{16}{*}{Thermo}                                                         & Gas Laws                & 5   \\
                                                                                 & Macro Heat, Work         & 5   \\
                                                                                 & Gibbs Free Energy & 4   \\
                                                                                 & Kinetic Theory of Gases                 & 2   \\
                                                                                 & Mixture                 & 2   \\
                                                                                 & Equilibrium             & 3   \\
                                                                                 & Translational Partition & 3   \\
                                                                                 & FRET Efficiency         & 1   \\
                                                                                 & Rate of Reaction        & 2   \\
                                                                                 & Transport Phenomena                & 2   \\
                                                                                 & Boltzman Distribution   & 2   \\
                                                                                 & Enthalpy
                                                                                    & 2    \\
                                                                                 & Entropy
                                                                                    & 4   \\   
                                                                                 & Electrolyte Solutions   & 2   \\
                                                                                 & Thermodynamics          & 3  \\
\midrule
All & & 122 \\
\bottomrule
\end{tabular}
\caption{Detailed distribution of our formula set.}
\label{tab: appendix formula dist}
\end{table}

\section{Details for Experiment}
\subsection{Experimental Setups}
\textbf{Few-shot Examples: }
we conduct all experiments under few-shot settings to better enable LLMs to solve physics problems. The examples are chosen from problems with detailed solutions in each dataset. Following SciBench, we use 3 examples for the thermo dataset and 4 examples for the other two datasets. 

\noindent
\textbf{Evaluation Criteria: }
the prediction is considered correct if it deviates from the correct answer by no more than 5\%. For system and CoT, we prompt the LLMs to output the predicted answer in "\textbackslash \textbackslash boxed\{\}". For other methods, the output Python code is executed which prints the predicted answer. If the output cannot be directly converted to a float, we capture the first number as the result.

\subsection{Prompt Templates for \Ours{}}
The prompt templates for three stages in \Ours{} are shown in Table~\ref{tab: our generation prompt}. In the formula retrieval stage, our approach first decides relevant subfields and then retrieves formulae from them.
\begin{table*}[t]
\centering
    \small
    \begin{tabular}{
        p{.16\textwidth}p{.8\textwidth}
        }
        \toprule
        \makecell[c]{\textbf{Prompt Templates for Physics Reasoner}} \\
        \midrule
        {Problem Analysis} &
        \makecell[{{>{\raggedright\arraybackslash}p{.8\textwidth}}}]{
            \texttt{You are an excellent student majoring in physics. Given a physics problem, please extract known variables from it in the format of Python code. The problem is intended to cover general topics within [FIELDS], sourced from a college-level textbook, and it requires analytical reasoning in physics.}\\
            The extracted variables should be well-named to represent the objects to which they belong and should be well-annotated with comments that provide sufficient explanations of the variables defined, including their units. Encase the Python code within triple backticks for clarity.\\ \\
            \texttt{\{ Example Problem\}} \\ 
            \texttt{\{ Example Analysis\}} \\ \\
            \texttt{\{ Problem\}}
        } \\
        \midrule
        {Subfield determination} &
        \makecell[{{>{\raggedright\arraybackslash}p{.8\textwidth}}}]{
            \texttt{You are an excellent student majoring in physics, given a physics problem, that could be related to several subfields of physics, including: [POTENTIAL SUBFIELDS], please determine and return relevant subject(s) from the provided list.}\\ \\ 
            \texttt{\{ Example Problem\}} \\ 
            \texttt{\{ Example Subfield(s)\}} \\ \\
            \texttt{\{ Problem\}}  
        } \\
        \midrule
        {Formula Retrieval} &
        \makecell[{{>{\raggedright\arraybackslash}p{.8\textwidth}}}]{
            \texttt{You are an excellent student majoring in physics, given a physics problem, please identify and retrieve relevant formulas. Annotate these relevant formulas in Python comments format and encase the Python code within triple backticks for clarity.}\\ \\
            \texttt{\{ Example Problem\}} \\ 
            \texttt{\{ Example Potential Formulas in Chosen Subfiled(s)\}} \\
            \texttt{\{ Example Retrieved Formulas\}} \\ \\
            \texttt{\{ Problem\}} \\
            \texttt{\{ Potential Formulas in Chosen Subfiled(s)\}}\\
        }\\
        \midrule
        {Guided Reasoning} & 
        \makecell[{{>{\raggedright\arraybackslash}p{.8\textwidth}}}]{
            \texttt{You are an excellent student majoring in physics, given a physics problem and its incomplete Python code, please think and complete the code. Print the target variable at the end if the problem has been solved. If there exists more than 1 target variable, print the first one. Remember to check whether the formulas are correctly used. And check whether the variable is printed with the required unit. Return the complete Python code and ensure it within triple backticks for clarity.}\\ \\
            \texttt{\{ Example Problem\}} \\ 
            \texttt{\{ Example Incomplete Code\}} \\
            \texttt{\{ Example Code\}} \\ \\
            \texttt{\{ Problem\}} \\
            \texttt{\{ Incomplete Code\}}\\
        }\\
        \bottomrule
    \end{tabular}
    \caption{Prompt templates for \Ours{}.}

    \label{tab: our generation prompt}
\end{table*}

\subsection{Prompt Templates for Baselines}
The prompt templates for the five baselines are shown in Table~\ref{tab: baseline prompt}.
For system, CoT, and PoT, we follow the settings in SciBench to ensure the effectiveness of comparisons. For the other two baselines, \textit{i.e.} self-correction and self-refine, we adapt their original prompts to fit the physics problem reasoning task.

\begin{table*}[t]
\centering
    \small
    \begin{tabular}{
        p{.16\textwidth}p{.8\textwidth}
        }
        \toprule
        \makecell[c]{\textbf{Prompt Templates for Baselines}} \\
        \midrule
        {Sys} &
        \makecell[{{>{\raggedright\arraybackslash}p{.8\textwidth}}}]{
            \texttt{Please provide a clear and step-by-step solution for a scientific problem in the categories of Chemistry, Physics, or Mathematics. The problem will specify the unit of measurement, which should not be included in the answer. Express the final answer as a decimal number with three digits after the decimal point. Conclude the answer by stating "The answer is therefore \textbackslash \textbackslash boxed\{[ANSWER]\}.}\\ \\
            \texttt{\{ Example Problem\}} \\ 
            \texttt{\{ Example Answer\}} \\ \\
            \texttt{\{ Problem\}}
        } \\
        \midrule
        {CoT} &
        \makecell[{{>{\raggedright\arraybackslash}p{.8\textwidth}}}]{
            \texttt{Please provide a clear and step-by-step solution for a scientific problem in the categories of Chemistry, Physics, or Mathematics. The problem will specify the unit of measurement, which should not be included in the answer. Express the final answer as a decimal number with three digits after the decimal point. Conclude the answer by stating "The answer is therefore \textbackslash \textbackslash boxed\{[ANSWER]\}.}\\ \\ 
            \texttt{\{ Example Problem\}} \\ 
            \texttt{\{ Example Solution\}} \\ \\
            \texttt{\{ Problem\}}  
        } \\
        \midrule
        {PoT} &
        \makecell[{{>{\raggedright\arraybackslash}p{.8\textwidth}}}]{
            \texttt{\# Please provide a clear and step-by-step solution for a scientific problem in the categories of Chemistry, Physics, or Mathematics. The problem will specify the unit of measurement. And potential related formulas are given. Please translate the solution steps into Python code and encase the Python code within triple backticks for clarity. \# }\\ \\
            \texttt{\{ Example Problem\}} \\ 
            \texttt{\{ Example Code\}} \\ \\
            \texttt{\{ Problem\}}
        }\\
        \midrule
        {Self-Correction (i) Feedback} & 
        \makecell[{{>{\raggedright\arraybackslash}p{.8\textwidth}}}]{
            \texttt{Review your previous answer and find problems with your answer.}\\ \\
            \texttt{\{Problem\}}\\
            \texttt{\{Code\}}
        }\\
        \midrule
        {Self-Correction (ii) Correction} & 
        \makecell[{{>{\raggedright\arraybackslash}p{.8\textwidth}}}]{
            \texttt{Based on the problems you found, improve your code. Please reiterate your code, with your final answer a single numerical number, in the form \textbackslash \textbackslash boxed\{answer\}. Encase the Python code within triple backticks for clarity.}\\ \\
            \texttt{\{Code\}}\\
            \texttt{\{Feedback\}}
        }\\
        \midrule
        {Self-Refine} & 
        \makecell[{{>{\raggedright\arraybackslash}p{.8\textwidth}}}]{
            \texttt{\{Example Code\}}\\
            \texttt{\# There is an error in the code above because of lack of understanding of the question. What is the error? To find the error, go through semantically complete blocks of the code, and check if everything looks good.}\\
            \texttt{\{Example Error\}}\\
            \texttt{\# Okay! Here is the rewrite:}\\
            \texttt{\{Example Refined code\}}\\ \\
            \texttt{\{Code\}}\\
            \texttt{\# There is an error in the code above because of lack of understanding of the question. What is the error? To find the error, go through semantically complete blocks of the code, and check if everything looks good.}\\
        }\\
        \bottomrule
    \end{tabular}
    \caption{Prompt templates for baselines.}

    \label{tab: baseline prompt}
\end{table*}

\subsection{Case Study}
\begin{figure*}[t]
  \centering
  \includegraphics[width=\linewidth]{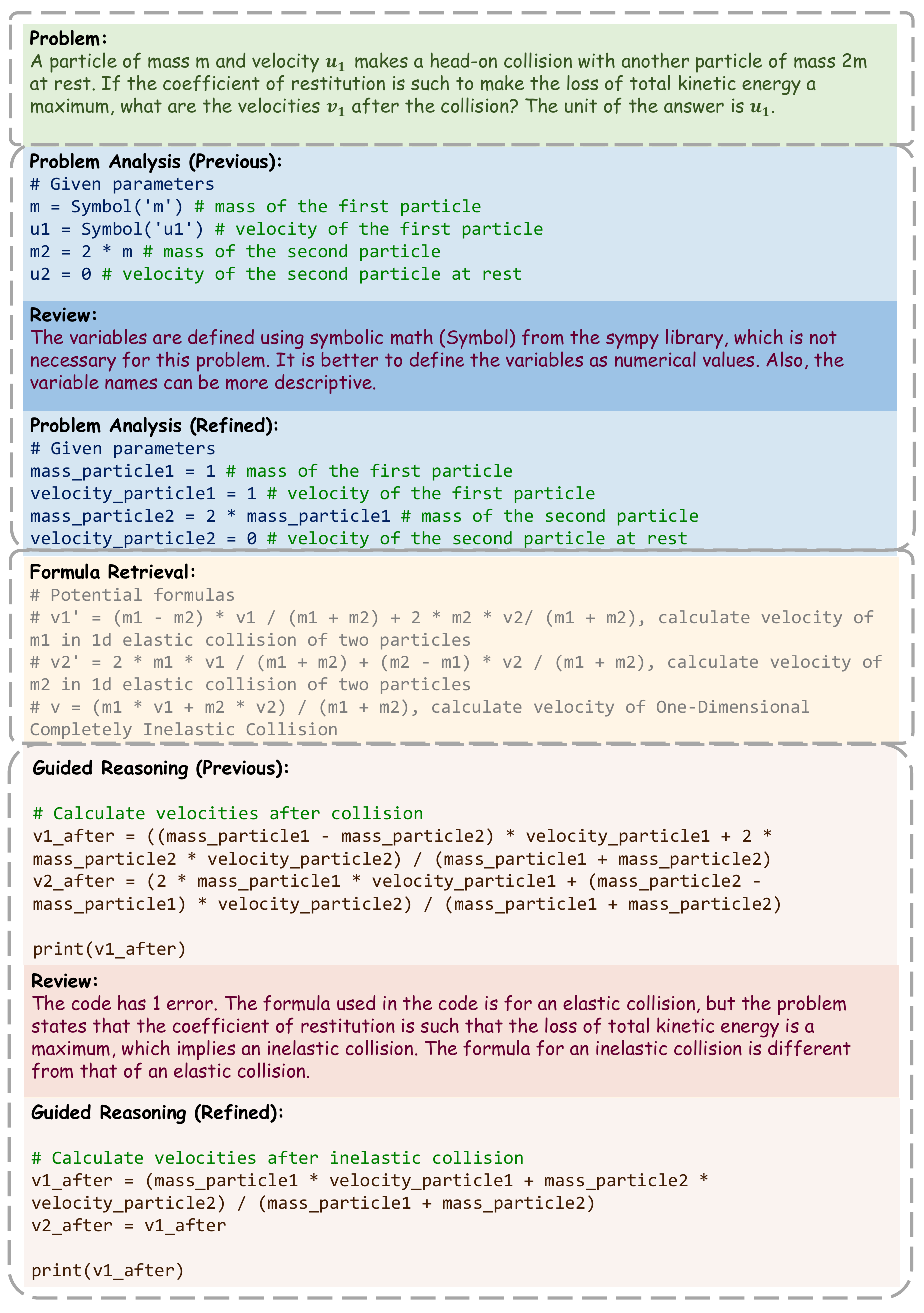}
  \caption{Case study using \Ours{} method.
  }
  \label{fig:appendix example}
\end{figure*}

We show a representative example by \Ours{} method in Fig.~\ref{fig:appendix example}.
In this problem, it successfully comprehends the problem, retrieves proper formulae, and effectively corrects mistakes with checklists. 

\section{Further Experiments}
\revise{
\subsection{Inclusion of Additional Datasets}
To confirm \Ours{}'s generalization ability, we also conduct experiments on a more challenging dataset, \textit{i.e.}, TheoremQA. The accuracy results (\%) with gpt-3.5-turbo are represented in Table~\ref{tab: theoremqa}. Following the TheoremQA approach, we compare the predicted answers with the correct answers, allowing a relative tolerance of 4\%. The temperature is set to zero to eliminate randomness.
\begin{table}[H]
\small
\centering
\begin{tabular}{lc}
\toprule
\textbf{Methods}         & \textbf{Accuracy} \\
\midrule
CoT                      & 7.6      \\
PoT                      & 14.5     \\
PoT+Self-Refine          & 21.4     \\
PoT+Self-Correction      & 22.1     \\
CREATOR                  & 16.8     \\
Chameleon                & 22.9     \\
\midrule
\textbf{\Ours{}} & 26.7    \\
\bottomrule
\end{tabular}
\caption{Additional results on TheoremQA.}
\label{tab: theoremqa}
\end{table}
Compared to other baselines, \Ours{} enhances accuracy from 22.9\% to 26.7\%, resulting in an average improvement of 3.8\%. The gains observed on both Scibench and TheoremQA demonstrate that \Ours{} outperforms other methods across various physics datasets. Additionally, \Ours{} exhibits strong generalization, boosting reasoning ability for challenging and infrequent physics problems.
}

\revise{
\subsection{Direct Comparison with Other Self-Revision Techniques}
We have also conducted experiments comparing our approach with other self-revision techniques, namely self-correction and self-refine, using the \textit{fund} dataset. The accuracy results (\%) are shown in Table~\ref{tab: appendix_comparison_self_revision}.
\begin{table}[H]
\small
\centering
\resizebox{\linewidth}{!}{
\begin{tabular}{lc}
\toprule
\textbf{Methods}         & \textbf{Accuracy} \\
\midrule
PoT + Self-Correction      & 30.1    \\
PoT + Self-Correction + Formula Set      & 31.5    \\
PoT + Self-Refine          & 30.1     \\
PoT + Self-Correction + Formula Set      & 32.9     \\
\midrule
\textbf{\Ours{}} & 38.4    \\
\bottomrule
\end{tabular}
}
\caption{Comparison with other self-revision baselines with our Formula Set on the fund dataset.}
\label{tab: appendix_comparison_self_revision}
\end{table}
The results indicate that Checklists are more effective than other self-revision techniques. Checklists are carefully designed to assist LLMs in self-improvement by enabling more accurate application of physics knowledge. In contrast, other self-revision techniques may overlook common mistakes in physics problems, resulting in inaccurate correction.
}

\end{document}